\documentclass[conference]{IEEEtran}
\IEEEoverridecommandlockouts
\usepackage{cite}
\usepackage{amsmath,amssymb,amsfonts}
\usepackage{algorithmic}
\usepackage{graphicx}
\usepackage{textcomp}
\usepackage{xcolor}
\usepackage{subfigure}
\usepackage{subfig}
\usepackage{graphicx}
\usepackage[utf8]{inputenc}
\usepackage{longtable}
\usepackage{booktabs}
\usepackage{textgreek}
\usepackage[font=small]{caption}
\usepackage{mathptmx}
\usepackage[ruled,vlined,linesnumbered]{algorithm2e}

\SetCommentSty{mycommfont}
\def\BibTeX{{\rm B\kern-.05em{\sc i\kern-.025em b}\kern-.08em
    T\kern-.1667em\lower.7ex\hbox{E}\kern-.125emX}}
    
\begin{document}

\title{Security Analysis of Capsule Network Inference using Horizontal Collaboration\\
}
\author{\IEEEauthorblockN{Adewale Adeyemo$^*$, Faiq Khalid$^\dag$,  Tolulope Odetola$^*$, Syed Rafay Hasan$^*$  }
\vspace{1mm}
\IEEEauthorblockA{\textit{$^*$Department of Electrical and Computer Engineering,  Tennessee Technological University,  Cookeville, TN 38505, USA}\\
}

\IEEEauthorblockA{\textit{$^\dag$Department of Computer Engineering, Vienna University of Technology, 1040 Wien, Austria}}\\

\vspace{-15mm}
}
\maketitle
\begin{abstract}

The traditional convolution neural networks (CNN) have several drawbacks like the “Picasso effect” and the loss of information by the pooling layer. The Capsule network (CapsNet) was proposed to address these challenges because its architecture can encode and preserve the spatial orientation of input images. Similar to traditional CNNs, CapsNet is also vulnerable to several malicious attacks, as studied by several researchers in the literature. However, most of these studies focus on single-device-based inference, but horizontally collaborative inference in state-of-the-art systems, like intelligent edge services in self-driving cars, voice controllable systems, and drones, nullify most of these analyses. Horizontal collaboration implies partitioning the trained CNN models or CNN tasks to multiple end devices or edge nodes. Therefore, it is imperative to examine the robustness of the CapsNet against malicious attacks when deployed in horizontally collaborative environments. Towards this, we examine the robustness of the CapsNet when subjected to noise-based inference attacks in a horizontal collaborative environment. In this analysis, we perturbed the feature maps of the different layers of four DNN models, i.e., CapsNet, mini-VGGNet, LeNet, and an in-house designed CNN (ConvNet) with the same number of parameters as CapsNet, using two types of noised-based attacks, i.e., Gaussian Noise Attack and FGSM noise attack. The experimental results show that similar to the traditional CNNs, depending upon the attacker’s access to the DNN layer, the classification accuracy of the CapsNet drops significantly. For example, when Gaussian Noise Attack classification is performed at the Digit-cap layer of the CapsNet, the maximum classification accuracy drop is approximately 97\%. Similarly, the maximum classification accuracy drop is 90.1\% when an FGSM noise attack is performed at the Conv layer of the CapsNet.  

\end{abstract}



\section{Introduction}
Convolution Neural Networks has been successful in many fields which include computer vision and Natural Language Processing \cite{marchisio2019capsattacks,wang2020convergence,bhandare2018designing,dahl2013large,pei2017deepxplore,hailesellasie2019vaws}. Due to the success recorded by Convolution Neural Networks (CNN) in classification tasks, they are being used in security-sensitive areas \cite{dahl2013large}, \cite{pei2017deepxplore}. 
However, the conventional CNNs have their shortcomings; they are unable to preserve the spatial relationship between objects, they also lose a lot of relevant information in the pooling layer. The challenge posed by the pooling layer is less trivial when classifying the whole image, but extremely challenging when performing image segmentation or object detection which requires preservation of pose (position, orientation, size, hue, albedo, etc.)\cite{michels2019vulnerability}.

To address the shortcomings of the pooling layer, Hinton et al.\cite{sabour2017dynamic} proposed a different type of network known as ‘Capsule Network (CapsNet)’ which relies on ‘capsules’ as a replacement for neurons in Deep Learning \cite{sabour2017dynamic}. 
CapsNet offers the advantage of preserving the position and pose information of features, high accuracy on the MNIST dataset \cite {lecun1998mnist}, and showed improved performance with small training dataset. Though with these advantages, CapsNet has its shortcoming which includes: its performance on larger images, slow training duration due to the inner loop of the dynamic routing algorithm and “crowding” problems (inability to recognize two identical objects) \cite {doerig2020capsule}. 
Several other researchers have previously investigated the adversarial robustness of CapsNet in a white box scenario and also proffered innovative methodologies to improve defense mechanisms to proposed adversarial attacks\cite{marchisio2019capsattacks,frosst1811darccc,peer2018training,sabour2017dynamic}. 

 To accommodate the high computation and memory requirement for the deployment of the CapsNet on hardware accelerators like FPGAs, horizontal collaborative inference \cite{mao2017local}, \cite{wang2020convergence}, \cite{mao2017modnn} could be employed. To achieve short-time-to-market, the deployment of the CNN model can be outsourced to third party designers (3PIP) \cite{odetola2021sowaf}. These 3PIP designers may be malicious. During horizontal collaborative inference, different partitions of the CNN layers are assigned to different 3PIPs, where no one 3PIP has access to the full CNN architecture. The 3PIP only has access to the allotted  parameters of the CNN layers and their respective output feature maps. In this paper, we investigate the vulnerability of the CapsNet network where the layers of the capsule network can be attacked by randomly introducing perturbations (adversarial attacks) in selected cycles or to the feature maps of model layers, forcing the model to wrongly classify a target class \cite{szegedy2013intriguing}.

\begin{figure}[h]
\centerline{\includegraphics[width=3.6in]{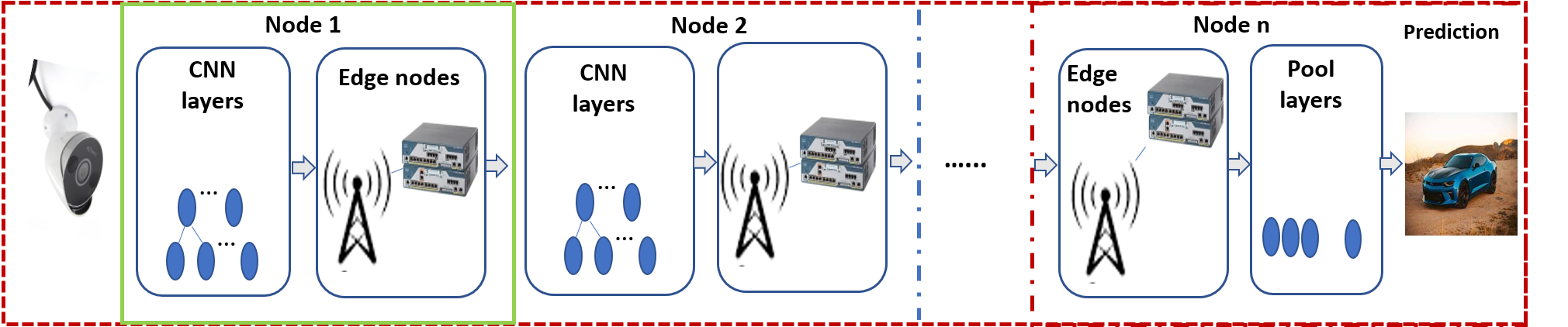}}
\caption{Different layers of a CNN model in different nodes}
\label{fig1}
 \vspace{-2mm}
\end{figure}
\textbf{Research Challenge:} In examining how robust/vulnerable the layers of CapsNet are in the face of a noised based attack. The question remains if capsule networks is robust in an horizontal collaborative environment? 

\textbf{Novel Contributions}
To address the aforementioned question, this paper focuses on analyzing the robustness of Capsule networks when subjected to noised-based attacks. This would involve applying perturbations to target layers in the CapsNet network. These perturbations would affect the values of the feature maps (output of the layer of a CNN model)  and impact the accuracy of the model. The results were benchmarked with three (3) regular CNNs (LeNet, Mini-Vgg, and ConvNet). We created a CNN model (ConvNet) to have the same number of parameters as CapsNet which forms a basis for comparing both networks. The training parameters are shown in Table \ref{tabb1}. We further analyzed the robustness of CapsNet against the simulated noised based attacks and benchmarked the performance against the three regular CNNs. 

\begin{table}[]
\centering
\caption{Training parameters of classification models.}

\resizebox{1\linewidth}{!}{
    \begin{tabular}{|c|c|c|c|c|}
    \hline
    \textbf{Parameters} & \textbf{\textit{CapsNet}}& \textbf{\textit{Lenet}}& \textbf{\textit{MiniVgg}} & \textbf{\textit{ConvNet}}\\
    \hline
    Optimization Method& Adam & SGD & SGD & SGD \\
    \hline
    Learning rate & 0.001& 0.01& 0.01& 0.01  \\
    \hline
    Epochs & 20 & 20 & 20 & 20 \\
    \hline
    No of Parameters (in Millions) & 8.2 & 1.25 & 1.7 & 8.2 \\
    \hline
    Baseline Inference Accuracy (percentage) & 99.6 & 99.1 & 99.1 & 98.2 \\
    \hline
    \end{tabular}}
\label{tabb1}
\vspace{-4mm}
\end{table}

\begin{figure}[h]
\centerline{\includegraphics[width=3.5in]{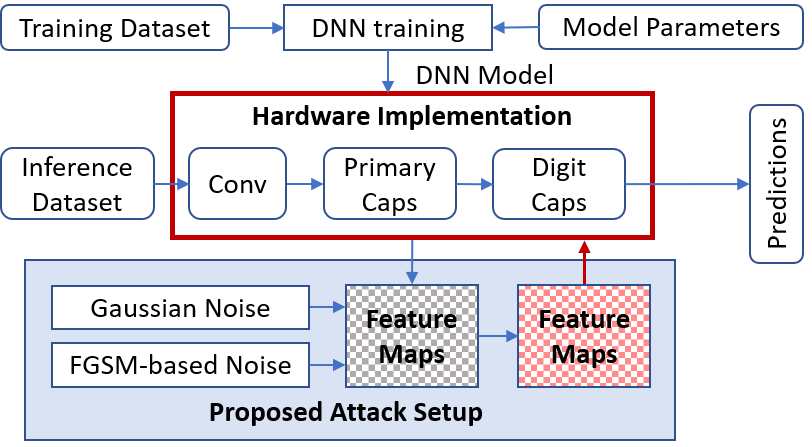}}
\caption{Novel framework to perform noised based attack on CapsNet by adding perturbations to individual layers of the Network}
\label{figg}
\end{figure}

To test the robustness of the CapsNet, as shown in Fig. \ref{figg}, we applied two types of perturbations to the layers within the networks. The perturbations would change the values of the feature maps of target layers, which would in turn affect accuracy of the model.

The remainder of the paper is structured as follows: Section II describes the threat model, Section III describes the attack methodology, section IV describes the results and Section V concludes our paper.

\section{Threat Model}
\begin{figure}[]
\centerline{\includegraphics[width=3.6in]{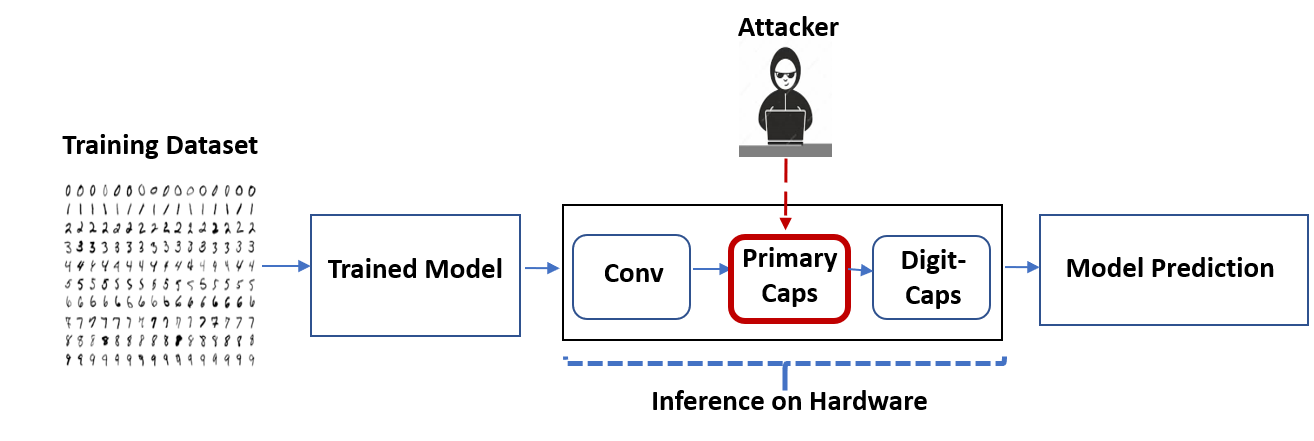}}
\caption{Targeted threat model: The 3PIP is assumed to have access to only some layers of the model}
\vspace{-5mm}
\label{fig_threat}
\end{figure}

This work proposes a graybox noised based attack that assumes the 3PIP designer is malicious with access to only the full CNN architecture or some of the CNN layers (as in the case of distributed network where horizontal collaboration is utilized) as shown in Fig. \ref{fig_threat}. We assume that the 3PIP has no access to training and testing data samples of the original CNN or CapsNet. It is also assumed that the attacker is provided with some validation dataset in conformity with industry standards \cite{odetola20192l}.

\section{Test of Robustness: Attack Methodology}
\begin{figure*}
    \centering
    \subfigure[Addition of noise to feature maps of target layers in CapsNet]{
    \includegraphics[width=0.45\linewidth, height= 1.2in]{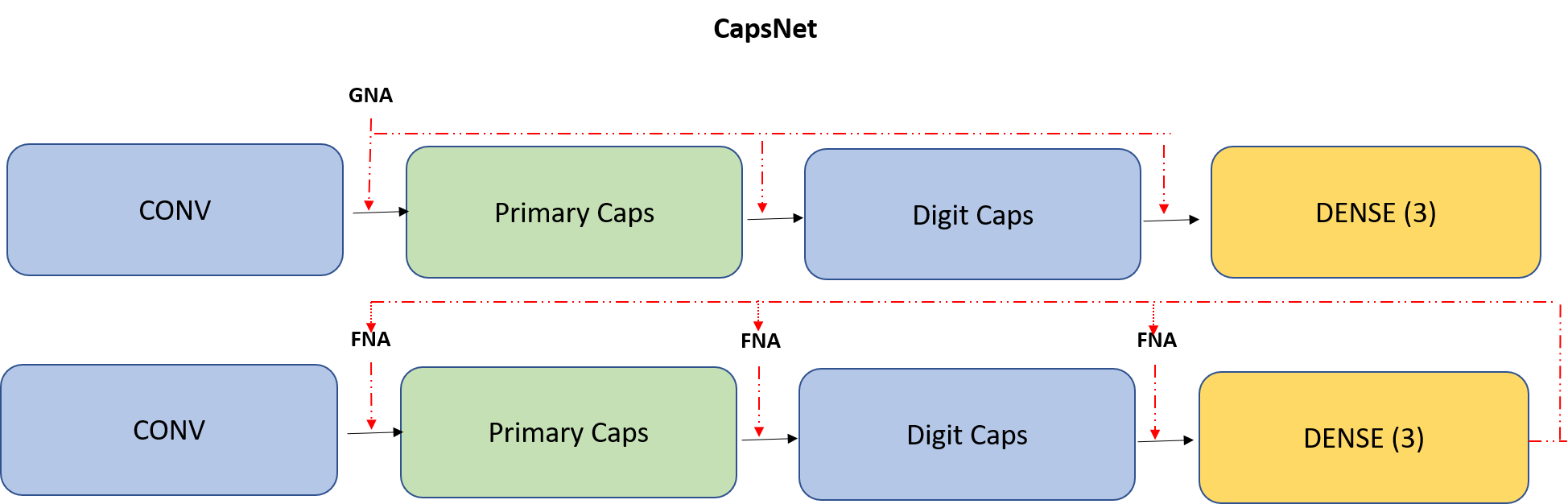} 
        }
    \subfigure[Addition of noise to feature maps of target layers in LeNet]{
    \includegraphics[width=0.45\linewidth, height= 1.1in ]{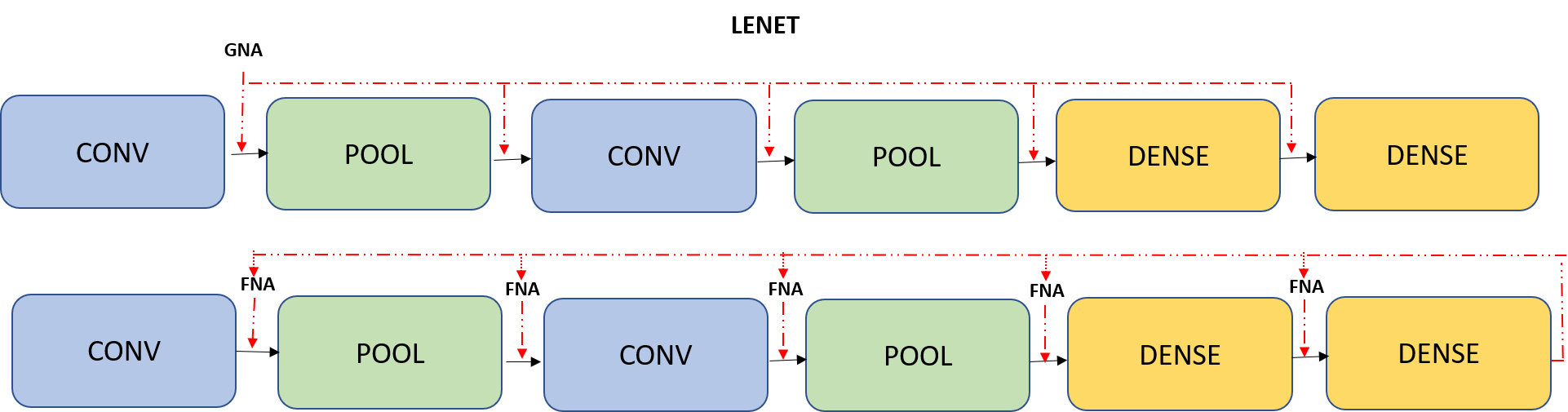} 
        }
    \subfigure[Addition of noise to feature maps of target layers in ConvNet]{
    \includegraphics[width=0.45\linewidth, height= 1.0in]{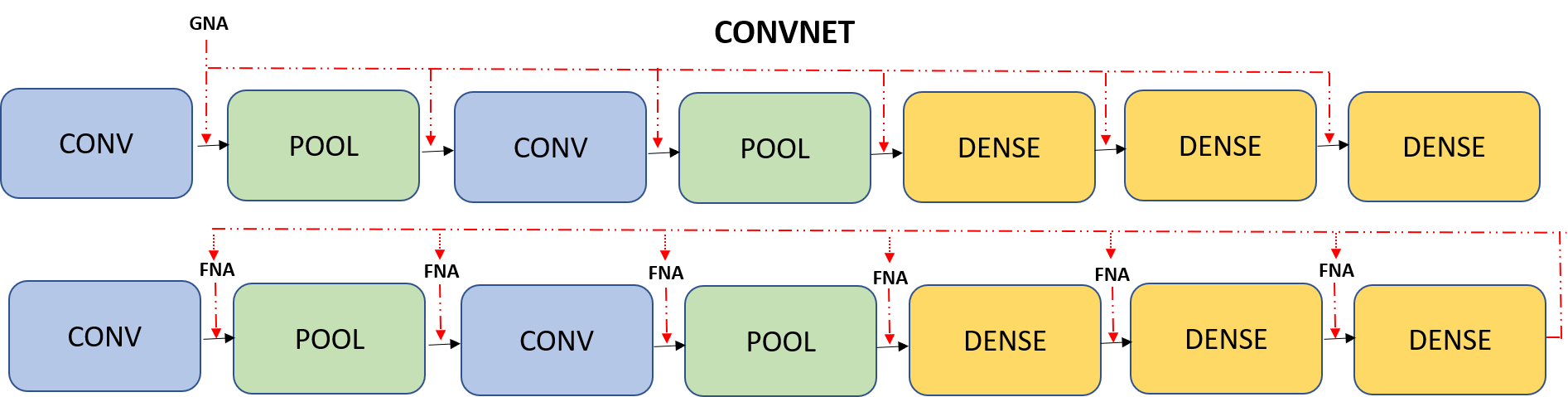} 
        }
    \subfigure[Addition of noise to feature maps of target layers in Mini-Vgg]{\includegraphics[width=0.45\linewidth, height= 1in]{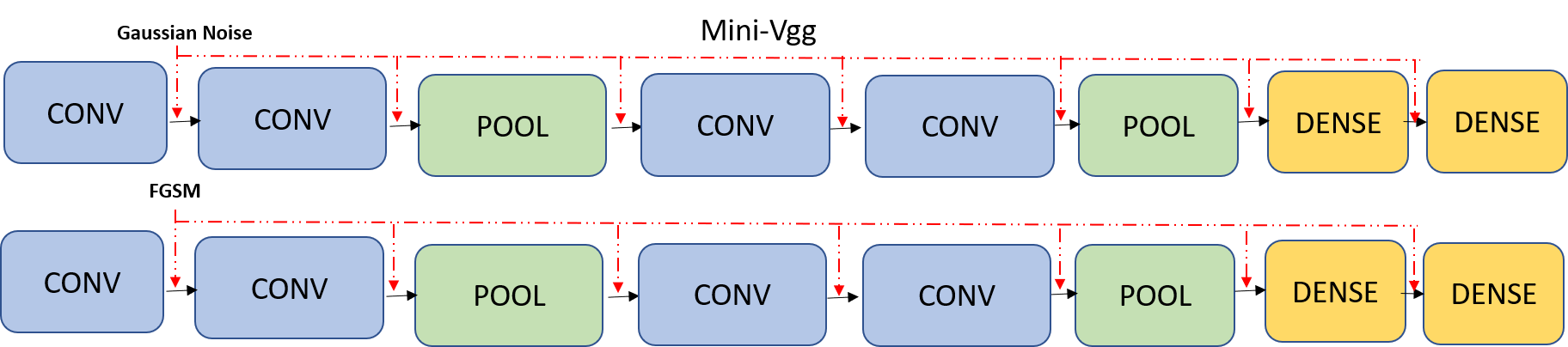}}
    
    \caption{Gaussian noise and FGSM induced noise being introduce to target layers under a noised based assumption that the 3PIP has access to specific layers of the model. This is expected to alter the values of feature maps in the target layers and ultimately affect classification accuracy. } 
    \label{ff}
 \vspace{-4mm}
\end{figure*}

 We utilized two types of noised based attacks with the first attack involving the application of Gaussian noise to a targeted layer without the knowledge of other model layer parameters. Here, we assumed the malicious person has access to only the target layer. For the second type of attack, we implemented the traditional FGSM under the assumption that the layers after the target layer is known to the 3PIP. The two noise based attacks utilized in exploring the vulnerability of CapsNet are discussed below:
\begin{itemize}

    \item Guassian Noise Attack (GNA): To carry out this attack, we assume that the 3PIP has access to only a target layer. A target layer is being selected (as shown in line 1 in Algorithm 1), Stochastically generated Gaussian noise ranging between 0 and 1 (with a standard deviation of 0.5), was added to varying number of elements (as shown in line 2 in Algorithm 1) in the different channels of the feature map of the target layer. The addition of Gaussian noise causes selective alteration in the values of the feature map of the target layer  (as shown in line 4-5 in Algorithm 1). The model is saved as shown in line 6 of Algorithm 1. Inference is performed on the model using the validation dataset to obtain the accuracy of model as shown in line 7.

    \item FGSM Noise Attack (FNA): To achieve this attack, we assume that the 3PIP has access to other layers after the target layer but not the layers before the target layer. Fast Gradient Signed Method (FGSM)\cite{xiao2018generating} was computed using the signed gradient of the loss of the model and multiplied with small value of \textepsilon\ to keep the perturbations imperceptible to the human eye in a bid to maintain stealthiness (as shown in line 2 Algorithm 2). The perturbations were added to some elements in the feature map of the target model layer rather than input images as shown in Fig. \ref{ff} and line 5 of Algorithm 2. The addition of perturbations causes selective alteration in the values of the feature map of the target layer  (as shown in line 5 of Algorithm 2). The model is saved as shown in line 6 of Algorithm 2. Inference is performed on the model using the validation dataset to obtain the accuracy of model as shown in line 7-8.
    
\end{itemize}
     After perturbations are added to the target layer, the model parameters still remains exactly the same except for the feature maps of the target layer.

\begin{algorithm}[!t]

 layer\_i = target\_layer; \tcp{i = targeted layer number}
 feature\_maps = nn\% of element in layer\_i; \tcp{ where nn = [25, 50, 75]}
 $x= [x_1....x_n] \in [0,1]\ $;\\ 
\SetAlgoLined
 \For {feature\_maps in layer\_i,}{
    $ feature\_maps^{'}= x * feature\_maps\;$\\
    $ model= save (feature\_maps)\ $;
    $ accuracy= test (model, validation\_set)\ $;
    $ return (accuracy) $
 }
 \caption{Gaussian noise-based perturbation in horizontally collaborative inference of DNNs}

\end{algorithm}



\begin{algorithm}[!t]
layer\_i=target\_layer; \tcp{i = layer number or target layer}
feature\_maps = nn\% of element in layer\_i; \tcp{where nn = [25, 50, 75]}
$ perturbations=  \epsilon *sign(\nabla_x J(\theta, feature\_maps,y));$\\ 
$where J= Loss,   \theta= model parameters,  y = labels; $\\
\SetAlgoLined
 \For {feature\_maps in layer\_i,}{
    $ feature\_maps^{'}= feature\_maps + perturbations \; $\\
    $ model= save (feature\_maps)\ $ ;
    $ accuracy= test (model, validation\_set)\ $;
    $ return (accuracy)$
 }

 \caption{FGSM-based noise perturbation in horizontally collaborative inference of DNNs}
 
\end{algorithm}

\section{Results}

To illustrate the noised based attacks, we trained the CapsNet and the regular convolution networks (Lenet, Mini-Vgg \& Convnet) on MNIST dataset using parameters shown in Table \ref{tabb1} above. Two types of attacks were carried out according to the threat model. The first attack (GNA) required only the knowledge of the target layer or the layer assigned to the 3PIP. In GNA, randomly generated noise with a standard deviation of 0.5 was clipped to \textit{selected} elements in the output feature maps of the target layer. These added values to the output feature maps introduce changes to the CapsNet model. The second attack (FNA) requires the 3PIP to have knowledge of the classification output of the original model. In FNA, perturbations are generated using FGSM and clipped to some elements of the output feature maps of the target layer. 

As shown in Fig. \ref{figg}, inference is performed using validation dataset and we observe that the effect of adding Gaussian perturbations has little effect on classification accuracy when perturbations was added to the first three (3) layers. However, the 6th layer in the Mini-Vgg architecture gave an accuracy of 99\% which is assumed to be random. We further observed that there was a decline in accuracy when FNA was added to target layers. CapsNet had the least performance in all of the four models as shown in Table \ref{tabb3}. In summary, we observed that the proposed methodology of attacks applicable to adversarial attacks that can be modified to exploit the vulnerability in DNNs during the horizontally collaborative inference.

\begin{table*}[!t]
\centering
\caption{Table showing comparison of Model Performances when subjected to GNA Attack, i.e., classification accuracy compromise in percentages of CapsNet, LeNet, Mini-Vgg and ConvNet respectively}
\small\addtolength{\tabcolsep}{11pt}
\begin{tabular}{|c|c|c|c|c|c|c|c|l|}
\hline
\multicolumn{9}{|c|}{\textbf{CapsNet}}                                      \\ \hline
\multicolumn{2}{|c|}{\textbf{Model Layers}}                 & \multicolumn{2}{c|}{\textbf{Conv}} & \multicolumn{2}{c|}{\textbf{Primary Caps}} & \multicolumn{3}{c|}{\textbf{Digit Caps}}                       \\ \hline
\multicolumn{2}{|c|}{25\% of feature maps}                            & \multicolumn{2}{c|}{99\%}            & \multicolumn{2}{c|}{98.4\%}                  & \multicolumn{3}{c|}{4\%}                                    \\ \hline
\multicolumn{2}{|c|}{50\% of feature maps}                            & \multicolumn{2}{c|}{98.8\%}          & \multicolumn{2}{c|}{98.9\%}                  & \multicolumn{3}{c|}{4\%}                                    \\ \hline
\multicolumn{2}{|c|}{75\% of feature maps}                            & \multicolumn{2}{c|}{92.3\%}          & \multicolumn{2}{c|}{93.38\%}                 & \multicolumn{3}{c|}{2\%}                                    \\ \hline
\multicolumn{9}{|c|}{\textbf{LeNet}}                                                                                                                                                                               \\ \hline
\multicolumn{1}{|l|}{\textbf{Model Layers}} & \textbf{Conv} & \textbf{Pool}    & \textbf{Conv}   & \multicolumn{2}{c|}{\textbf{Pool}}         & \multicolumn{3}{c|}{\textbf{Dense}}                       \\ \hline
25\% of feature maps                                  & 99\%            & 99\%               & 99\%              & \multicolumn{2}{c|}{99\%}                    & \multicolumn{3}{c|}{5.1\%}                                  \\ \hline
50\% of feature maps                                  & 99\%            & 98\%               & 99\%              & \multicolumn{2}{c|}{75\%}                    & \multicolumn{3}{c|}{5.5\%}                                  \\ \hline
75\% of feature maps                                  & 99\%            & 98\%               & 98\%              & \multicolumn{2}{c|}{50\%}                    & \multicolumn{3}{c|}{4.8\%}                                  \\ \hline
\multicolumn{9}{|c|}{\textbf{Mini-VGG}}                                                                                                                                                                    \\ \hline
\textbf{Model}                              & \textbf{Conv} & \textbf{Conv}    & \textbf{Pool}   & \textbf{Conv}        & \textbf{Conv}       & \textbf{Pool}       & \multicolumn{2}{c|}{\textbf{Dense}} \\ \hline
25\% of feature maps                                  & 99\%            & 97.2\%             & 99.5\%            & 71.2\%                 & 99\%                  & 99\%                  & \multicolumn{2}{c|}{24\%}             \\ \hline
50\% of feature maps                                  & 98.5\%          & 98.4\%             & 99\%              & 71.9\%                 & 75\%                  & 99\%                  & \multicolumn{2}{c|}{20\%}             \\ \hline
75\% of feature maps                                  & 96.7\%          & 97\%               & 98\%              & 71.2\%                 & 50\%                  & 99\%                  & \multicolumn{2}{c|}{27\%}             \\ \hline
\multicolumn{9}{|c|}{\textbf{ConvNet}}                                                                                                                                                                    \\ \hline
\textbf{Model}                              & \textbf{Conv} & \textbf{Pool}    & \textbf{Conv}   & \textbf{Pool}        & \multicolumn{2}{c|}{\textbf{Conv}}        & \multicolumn{2}{c|}{\textbf{Dense}} \\ \hline
25\% of feature maps                                 & 98\%             & 96\%                & 89\%               & 90\%                    & \multicolumn{2}{c|}{59.7\%}                    & \multicolumn{2}{c|}{8.9\%}              \\ \hline
50\% of feature maps                                  & 97\%            & 96\%                & 82\%               & 71\%                    & \multicolumn{2}{c|}{42\%}                    & \multicolumn{2}{c|}{7.4\%}              \\ \hline
75\% of feature maps                                 & 97\&             & 95\%                & 86\%               & 82\%                    & \multicolumn{2}{c|}{40\&}                    & \multicolumn{2}{c|}{4\%}              \\ \hline

\end{tabular}

\label{tabb2}
\end{table*}


\begin{table*}[]
\centering
\caption{Table showing comparison of Model Performances when subjected to FNA Attack, i.e., classification accuracy compromise in percentages of CapsNet, LeNet, Mini-Vgg and ConvNet respectively}
\small\addtolength{\tabcolsep}{11pt}
\begin{tabular}{|c|c|c|c|c|c|c|c|}
\hline
                                                     
\multicolumn{8}{|c|}{\textbf{CapsNet}}                                                                                 \\ \hline
\textbf{Layer Considered} & \multicolumn{2}{c|}{\textbf{Conv}} & \multicolumn{3}{c|}{\textbf{Primary Caps}}      & \multicolumn{2}{c|}{\textbf{Digit Caps}}  \\ \hline
Accuracy         & \multicolumn{2}{c|}{8.9\%}  & \multicolumn{3}{c|}{49.7\%}             & \multicolumn{2}{c|}{18.9\%}       \\ \hline
\multicolumn{8}{|c|}{\textbf{LeNet}}                                                                                   \\ \hline
Layer Considered & \multicolumn{2}{c|}{\textbf{Conv}} & \textbf{Pool}      & \multicolumn{2}{c|}{\textbf{Conv}} & \textbf{Pool}           & \textbf{Dense}          \\ \hline
Accuracy         & \multicolumn{2}{c|}{89.1\%} & 81\%        & \multicolumn{2}{c|}{85.5\%} & 78.8\%           & 75.4\%           \\ \hline
\multicolumn{8}{|c|}{\textbf{Mini-VGG}}                                                                                \\ \hline
\textbf{Layer Considered} & \textbf{Conv}        & \textbf{Conv}        & \textbf{Pool}      & \textbf{Conv}        & \textbf{Conv}        & \textbf{Pool}           & \textbf{Dense}          \\ \hline
Accuracy         & 73.8\%        & 77.4\%        & 99\%        & 77.2\%        & 76.8\%        & 97.6\%           & 86\%             \\ \hline
\multicolumn{8}{|c|}{\textbf{ConvNet}}                                                                                 \\ \hline
\textbf{Layer Considered} & \textbf{Conv}        & \textbf{Pool}        & \textbf{Conv}      & \textbf{Pool}        & \textbf{Conv}        & \textbf{Dense}          & \textbf{Dense}          \\ \hline
Accuracy         & 82\%          & 89\%          & 75\%        & 78.8\%        & 11.9\%        & 9\%              & 1\%              \\ \hline

\end{tabular}
\label{tabb3}
\end{table*}

\section{Conclusion and Future Work}
In this paper, we examined the robustness of the Capsule Networks when subjected to noised based attacks inference attacks in a horizontal collaborative environment. In this work, the threat model assumes that an attacker is malicious and does not have access to the full CapsNet architecture. Two types of attacks were applied (GNA and FNA). We applied perturbations to different layers of the CapsNet model and observed the effect on the classification accuracy of the model. We further bench-marked the performance of the CapsNet model under attack against three regular CNN models. Our experiment showed that CapsNet model achieved similar classification accuracy  as compared to the regular CNNs when attacked using GNA; when Gaussian Noise Attack classification is performed at the Digit-cap layer of the CapsNet, the maximum classification accuracy drop is approximately 97\%. Similarly, the maximum classification accuracy drop is 90.1\% when an FGSM noise attack is performed at the Conv layer of the CapsNet. To further expand on this work, the next phase is to come up with defence mechanisms against the attacks discussed. 
\vspace{-4mm}

\bibliographystyle{IEEEtran}
\bibliography{references.bib}

\begin{thebibliography}{10}
\providecommand{\url}[1]{#1}
\csname url@samestyle\endcsname
\providecommand{\newblock}{\relax}
\providecommand{\bibinfo}[2]{#2}
\providecommand{\BIBentrySTDinterwordspacing}{\spaceskip=0pt\relax}
\providecommand{\BIBentryALTinterwordstretchfactor}{4}
\providecommand{\BIBentryALTinterwordspacing}{\spaceskip=\fontdimen2\font plus
\BIBentryALTinterwordstretchfactor\fontdimen3\font minus
  \fontdimen4\font\relax}
\providecommand{\BIBforeignlanguage}[2]{{%
\expandafter\ifx\csname l@#1\endcsname\relax
\typeout{** WARNING: IEEEtran.bst: No hyphenation pattern has been}%
\typeout{** loaded for the language `#1'. Using the pattern for}%
\typeout{** the default language instead.}%
\else
\language=\csname l@#1\endcsname
\fi
#2}}
\providecommand{\BIBdecl}{\relax}
\BIBdecl

\bibitem{marchisio2019capsattacks}
A.~Marchisio, G.~Nanfa, F.~Khalid, M.~A. Hanif, M.~Martina, and M.~Shafique,
  ``Capsattacks: Robust and imperceptible adversarial attacks on capsule
  networks,'' \emph{arXiv preprint arXiv:1901.09878}, 2019.

\bibitem{wang2020convergence}
X.~Wang, Y.~Han, V.~C. Leung, D.~Niyato, X.~Yan, and X.~Chen, ``Convergence of
  edge computing and deep learning: A comprehensive survey,'' \emph{IEEE
  Communications Surveys \& Tutorials}, vol.~22, no.~2, pp. 869--904, 2020.

\bibitem{bhandare2018designing}
A.~Bhandare and D.~Kaur, ``Designing convolutional neural network architecture
  using genetic algorithms,'' in \emph{\textbf{ICAI}}, 2018, pp. 150--156.

\bibitem{dahl2013large}
G.~E. Dahl, J.~W. Stokes, L.~Deng, and D.~Yu, ``Large-scale malware
  classification using random projections and neural networks,'' in \emph{IEEE
  \textbf{ICASSP}}, 2013, pp. 3422--3426.

\bibitem{pei2017deepxplore}
K.~Pei, Y.~Cao, J.~Yang, and S.~Jana, ``Deepxplore: Automated whitebox testing
  of deep learning systems,'' in \emph{\textbf{SOSP}}, 2017, pp. 1--18.

\bibitem{hailesellasie2019vaws}
M.~Hailesellasie, J.~Nelson, F.~Khalid, and S.~R. Hasan, ``Vaws: Vulnerability
  analysis of neural networks using weight sensitivity,'' in \emph{IEEE
  \textbf{MWSCAS}}, 2019, pp. 650--653.

\bibitem{michels2019vulnerability}
F.~Michels, T.~Uelwer, E.~Upschulte, and S.~Harmeling, ``On the vulnerability
  of capsule networks to adversarial attacks,'' \emph{arXiv preprint
  arXiv:1906.03612}, 2019.

\bibitem{sabour2017dynamic}
S.~Sabour, N.~Frosst, and G.~E. Hinton, ``Dynamic routing between capsules,''
  \emph{arXiv preprint arXiv:1710.09829}, 2017.

\bibitem{lecun1998mnist}
Y.~LeCun, ``The mnist database of handwritten digits,'' \emph{http://yann.
  lecun. com/exdb/mnist/}, 1998.

\bibitem{doerig2020capsule}
A.~Doerig, L.~Schmittwilken, B.~Sayim, M.~Manassi, and M.~H. Herzog, ``Capsule
  networks as recurrent models of grouping and segmentation,'' \emph{PLoS
  computational biology}, vol.~16, no.~7, p. e1008017, 2020.

\bibitem{frosst1811darccc}
N.~Frosst, S.~Sabour, and G.~Hinton, ``Darccc: Detecting adversaries by
  reconstruction from class conditional capsules. arxiv 2018,'' \emph{arXiv
  preprint arXiv:1811.06969}.

\bibitem{peer2018training}
D.~Peer, S.~Stabinger, and A.~Rodriguez-Sanchez, ``Training deep capsule
  networks,'' \emph{arXiv preprint arXiv:1812.09707}, 2018.

\bibitem{mao2017local}
J.~Mao, ``Local distributed mobile computing system for deep neural networks,''
  Ph.D. dissertation, University of Pittsburgh, 2017.

\bibitem{mao2017modnn}
J.~Mao, X.~Chen, K.~W. Nixon, C.~Krieger, and Y.~Chen, ``Modnn: Local
  distributed mobile computing system for deep neural network,'' in \emph{IEEE
  \textbf{DATE}}, 2017, pp. 1396--1401.

\bibitem{odetola2021sowaf}
T.~A. Odetola and S.~R. Hasan, ``Sowaf: Shuffling of weights and feature maps:
  A novel hardware intrinsic attack (hia) on convolutional neural network
  (cnn),'' \emph{arXiv preprint arXiv:2103.09327}, 2021.

\bibitem{szegedy2013intriguing}
C.~Szegedy, W.~Zaremba, I.~Sutskever, J.~Bruna, D.~Erhan, I.~Goodfellow, and
  R.~Fergus, ``Intriguing properties of neural networks,'' \emph{arXiv preprint
  arXiv:1312.6199}, 2013.

\bibitem{odetola20192l}
T.~A. Odetola, K.~M. Groves, and S.~R. Hasan, ``2l-3w: 2-level 3-way
  hardware-software co-verification for the mapping of deep learning
  architecture (dla) onto fpga boards,'' \emph{arXiv preprint
  arXiv:1911.05944}, 2019.

\bibitem{xiao2018generating}
C.~Xiao, B.~Li, J.-Y. Zhu, W.~He, M.~Liu, and D.~Song, ``Generating adversarial
  examples with adversarial networks,'' \emph{arXiv preprint arXiv:1801.02610},
  2018.

\end{thebibliography}
\end{document}